\documentclass[pdflatex,sn-basic]{sn-jnl}

\usepackage{graphicx}%
\usepackage{multirow}%
\usepackage{amsmath,amssymb,amsfonts}%
\usepackage{amsthm}%
\usepackage{mathrsfs}%
\usepackage[title]{appendix}%
\usepackage{xcolor}%
\usepackage{textcomp}%
\usepackage{manyfoot}%
\usepackage{booktabs}%
\usepackage{algorithm}%
\usepackage{algorithmicx}%
\usepackage{algpseudocode}%
\usepackage{listings}%

\usepackage[utf8]{inputenc} 
\usepackage[T1]{fontenc}    
\usepackage{hyperref}       
\usepackage{url}            
\usepackage{booktabs}       
\usepackage{amsfonts}       
\usepackage{nicefrac}       
\usepackage{microtype}      
\usepackage{xcolor}         
\usepackage{graphicx}
\usepackage{natbib}
\usepackage{subcaption}
\usepackage{wrapfig}
\usepackage{multirow}
\usepackage{setspace}

\usepackage{listings}
\usepackage{mathtools}

\begin{document}

\title[Article Title]{Intrinsically-Motivated Humans and Agents in Open-World Exploration}


\author*[1]{\fnm{Aly} \sur{Lidayan}}\email{dayan@berkeley.edu }
\equalcont{These authors contributed equally to this work.}

\author*[1]{\fnm{Yuqing} \sur{Du}}\email{yuqing\_du@berkeley.edu }
\equalcont{These authors contributed equally to this work.}

\author*[1,2]{\fnm{Eliza} \sur{Kosoy}}\email{eko@berkeley.edu}
\equalcont{These authors contributed equally to this work.}

\author[2]{\fnm{Maria} \sur{Rufova}}\email{mariarufova@berkeley.edu}

\author[1]{\fnm{Pieter} \sur{Abbeel}}\email{pabbeel@cs.berkeley.edu }

\author[1,2]{\fnm{Alison} \sur{Gopnik}}\email{gopnik@berkeley.edu}

\affil[1]{\orgdiv{BAIR}, \orgname{University of California Berkeley}, \orgaddress{\city{Berkeley}, \postcode{94706}, \state{CA}, \country{USA}}}


\affil[2]{\orgdiv{Psychology}, \orgname{University of California Berkeley}, \orgaddress{\city{Berkeley}, \postcode{94706}, \state{CA}, \country{USA}}}


\abstract{
\textit{What drives exploration?} Understanding intrinsic motivation is a long-standing challenge in both cognitive science and artificial intelligence; numerous objectives have been proposed and used to train agents, yet there remains a gap between human and agent exploration. We directly compare adults, children, and AI agents in a complex open-ended environment, Crafter~\citep{hafner2021benchmarking}, and study how common intrinsic objectives: Entropy, Information Gain, and Empowerment, relate to their behavior. We find that only Entropy and Empowerment are consistently positively correlated with human exploration progress, indicating that these objectives may better inform intrinsic reward design for agents. Furthermore, across agents and humans we observe that Entropy initially increases rapidly, then plateaus, while Empowerment increases continuously, suggesting that state diversity may provide more signal in early exploration, while advanced exploration should prioritize control. Finally, we find preliminary evidence that private speech utterances, and particularly goal verbalizations, may aid exploration in children. Our data is available at \url{https://github.com/alyd/humans_in_crafter_data}.
}




\maketitle

\section{Introduction}

\begin{figure}[ht]
\begin{center}
\vspace{-3ex}
\includegraphics[width=\textwidth]{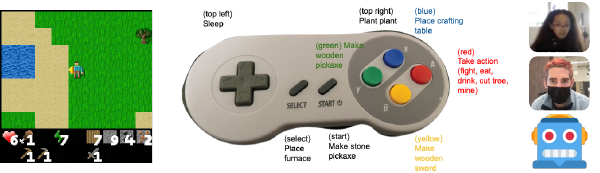}
\end{center}
\caption{\textbf{Left}: Example screen from Crafter \citep{hafner2021benchmarking}. The player is at the center of the viewing window; the yellow arrow shows which direction they are facing. Their health, food, water and energy status are at the bottom left, the raw materials they have collected are at the bottom right, and the tools built so far are in the bottom row. \textbf{Middle}: Actions available to the human participants and RL agents. \textbf{Right}: We compare behaviors of children, adults, and RL agents.} 
\label{fig:screen}
\end{figure}

Human beings explore new environments remarkably effectively, even in the total absence of external rewards \citep{matusch2020evaluating}.
Although adult humans have extensive experience and prior knowledge, even children explore intelligently~\citep{giron2023developmental,gopnik2020childhood}, suggesting that their behaviors might be particularly informative about the foundational cognitive processes that underpin the success of human exploration. 
How to endow artificial agents  with such exploration abilities is still an open question~\citep{hughes2024open}, despite their success in other domains \citep{kim2024openvla, silver2017mastering}. One approach to guiding exploration in Reinforcement Learning (RL) agents is to set up intrinsic rewards~\citep{aubret2019survey}. In contrast to extrinsic rewards used for learning specific tasks, such as winning chess games,
intrinsic rewards aim to be domain-agnostic to encourage open-ended exploration, often taking inspiration from information theory and human curiosity \citep{aubret2023information,schmidhuber1991possibility}.
However, it remains unclear which objectives truly underlie human curiosity and exploration~\citep{ten2022curiosity, pelz2020elaboration}, and only a few studies use concrete demonstrations of human exploration as a basis for agent exploration \citep{dubey2018investigating,bambach2018toddler,kosoy2020exploring,kosoy2022learning}.

Better understanding of human intrinsically-motivated exploration could yield insights for building more effective AI agents. To this end, we collect exploratory play data from adults, children, and RL agents in the same open-ended environment---Crafter~\citep{hafner2021benchmarking}, a Minecraft-like game which offers a wide breadth and depth of activities to explore. We train RL agents on a range of objectives, and analyze how well the exploration progress of humans and agents correlates with three popular information-theoretic objectives: \textit{Entropy}, \textit{Information Gain}, and \textit{Empowerment}. We find that human performance is consistently positively correlated with \textit{Entropy} and \textit{Empowerment}. Furthermore, in both humans and agents we observe that \textit{Entropy} initially increases rapidly before plateauing, while \textit{Empowerment} increases steadily, suggesting that state novelty may provide more signal in early exploration, whereas control may be a more effective objective later in exploration. We also transcribe human utterances during play, and find significant positive correlations between children's performance and their frequency of private speech utterances, particularly those verbalizing goals. This supports prior work in psychology which has suggested that private speech could play an important role in children’s creative problem-solving \citep{Lee_2011,vygotsky1987thinking}, and work in AI which finds that reasoning and setting goals explicitly in language improves problem-solving and exploration capabilities \citep{du2023guiding, zhang2023omni}. %

\textbf{Exploration in AI Agents.} 
Approaches to exploration range in complexity from occasionally taking random actions (\textit{e.g.}, $\epsilon$-greedy exploration) to automatic curriculum learning~\citep{portelas2020automatic} and goal-setting~\citep{colas2022autotelic}. Adding pseudo-rewards, known as intrinsic motivation or reward shaping, is a particularly versatile method for guiding exploration in reinforcement learning agents, which learn to maximize rewards (see~\citet{lidayanbamdp} for a survey). The universality of information-theoretic measures such as entropy, information gain, and empowerment make them popular inspiration for designing general-purpose rewards to motivate open-ended exploration~\citep{aubret2023information}. Entropy rewards are based on the intuition that exploratory behavior can be generated by visiting diverse states or taking diverse actions \citep{hazan2019provably, liu2021behavior, yarats2021reinforcement}. Information gain encourages gathering information about the environment and its dynamics~\citep{lindley1956measure}, e.g, by rewarding states or transitions that the agent cannot yet predict well~\citep{schmidhuber1991possibility, pathak2017curiosity, zhang2021noveld}, by leveraging Bayesian networks \citep{houthooft2016vime}, or network ensembles \citep{pathak2019self}. Empowerment, defined as the mutual information between an agent's actions and their outcomes, is maximized by taking diverse and controllable actions \citep{klyubin2005empowerment, klyubin2005all},  encouraging exploration that increases the agent's influence over the environment~\citep{salge2014changing,gregor2016variational}. 

However, even with sophisticated intrinsic objectives, RL agents often lag far behind human efficiency \citep{matusch2020evaluating,hafner2023mastering}. In some settings, intrinsic rewards can even harm exploration by unintentionally incentivizing degenerate behaviors, such as fixation on sources of random noise in the environment \citep{burda2018exploration}.
Another explanation for the gap between human and agent exploration abilities is the prior knowledge that humans have from life experience ~\citep{dubey2018investigating}.
That said, in RL systems, prior knowledge by itself is useless without an objective. Our work aims to understand which \textit{objectives} motivate human exploration, and how that can inform intrinsic reward design for agents.

\textbf{Exploration in Humans.} 
Dating back to \citet{Piaget1933}, developmental researchers have conceived of children as active and curious learners who are intrinsically motivated to explore the world in systematic and rational ways \citep{Schulz_2007,Cook,Legare_2012}; see \citet{Schulz_2012} for a review. As in the AI literature, a variety of intrinsic objectives have been proposed to underlie human curiosity and exploration \citep{ten2022curiosity, pelz2020elaboration}. 
Some evidence suggests that humans prefer tasks that maximize expected \textit{learning progress} \citep{baldassarre2014intrinsic, ten2021humans,oudeyer2007intrinsic}, suggesting that they are intrinsically motivated to improve their own performance. Information-theoretic objectives have likewise been proposed to drive human exploration. \citet{brandle2023empowerment} find that humans appeared to maximize empowerment when they explored the Little Alchemy game, preferring actions that allowed more potential possible actions later in the game, and \citet{melnikoff2022computational} find that maximizing empowerment was associated with the experience of "flow". There is also evidence of humans seeking experiences that maximize information gain, allowing them to represent the structure of the environment more accurately \citep{liquin2021developmental,ruggeri2024preschoolers,addyman2013local}. Finally, prior work has shown people are drawn to stimuli that appear more novel~\citep{taffoni2014development,berlyne1950novelty,smock1962children,poli2022contributions}, which closely relates to maximizing state entropy.

Computational studies of exploration in cognitive science thus far have generally been limited to simplistic and unrealistic environments, typically with only a couple different states or actions to choose from \citep{gershman2018deconstructing,baranes2014effects,ten2021humans}. Some more realistic studies have used 3D mazes \citep{kosoy2020exploring} or the Little Alchemy games \citep{brandle2023empowerment}. In this study we use a Minecraft-like environment which offers a particularly rich and open-ended set of possibilities. 
By comparing human and agent exploration in this shared environment, we hope to illuminate which objectives make humans so effective at exploration.

\textbf{Language and Exploration.}
Recent work in AI has found that agents can explore more effectively by reasoning with and generating goals using language \citep{wei2022chain, zhang2023omni,du2023guiding}.
Meanwhile, \citet{vygotsky1962thought} found 
verbalization (i.e., thinking aloud) to be especially frequent in children, particularly when encountering obstacles, and it has been found to aid problem solving by focusing attention on important features \citep{schunk1986verbalization,granato2020computational}, or assisting with coding and retention of information \citep{gidley2015contribution}. This motivated us to further probe the role of language in open-ended exploration, by analyzing private speech utterances from the human players.
\section{Results}
We conduct three main analyses on the recorded exploration data. First, we introduce scores for exploration in the Crafter environment and find that adults explore more effectively than RL agents, while intrinsically motivated agents are outperformed by agents trained on extrinsic rewards tailored to the underlying game structure, supporting the idea that there is room for improvement in intrinsic reward design. Next, we analyze to what extent human and agent exploratory behaviors optimize for commonly used information-theoretic intrinsic objectives: Entropy, Information Gain, and Empowerment. We find significant positive correlations between human exploration scores and both Entropy and Empowerment, but not Information Gain. Lastly, we study the relationship between private speech and exploration by analyzing human participant utterances. In children, though not adults, there is a strong positive correlation between the fraction of goal-based utterances and exploration score.

\subsection{Participants} Participants were allowed to play Crafter for up to 20 minutes. No direction was given about the game objectives and no score was visible in order to encourage intrinsically motivated, open-ended play. For AI agent comparison, we trained RL agents with three different objectives: two different intrinsic rewards based on entropy and information gain, and one extrinsic reward that was tailored to the structure of the game. 13 agents were trained per setting to reduce variance over 1 million time-steps. Finally, we used agents that sample actions uniformly at random as a baseline. See Section \ref{sec:methods} for details.

\subsection{Exploration Performance}

We design five scores measuring different aspects of exploration proficiency (Table \ref{tab:expldefs}). Overall Achievement Score (AS) measures the fraction of all possible achievements unlocked at least once over all played episodes, which we break down into Breadth and Depth to probe different potential exploration strategies. To investigate performance throughout, we also calculate the average number of achievements reached (Mean AS) and fraction of the world map physically visited in each episode. Note that a new map is procedurally generated at the start of each episode.
 
\begin{table}[ht]
    \centering
    \small
    \begin{tabular}{c|c}
         \toprule 
    \textbf{Exploration Score} & \textbf{Definition}  \\ \midrule
        Mean Achievement Score & \parbox{0.65\linewidth}{The average fraction of unique achievements reached per episode.}\\ 
        \midrule
        Map Coverage Score &\parbox{0.65\linewidth}{ The average fraction of the world map visited per episode.} \\
        \midrule
        Overall Achievement Score & \parbox{0.65\linewidth}{Fraction of unique achievements unlocked over all episodes.}\\ \midrule
        Overall Breadth Score & \parbox{0.65\linewidth}{Fraction of steps taken in a breadth-first traversal of the achievement tree (\textit{i.e.} only counting achievements up to and including the first incomplete tree level). }\\ \midrule
        Overall Depth Score & \parbox{0.65\linewidth}{ Normalized depth of the deepest achievement reached in the achievement tree.} \\   \bottomrule
    \end{tabular}
    \caption{Description of exploration scores proposed for our study.}
    \label{tab:expldefs}
\end{table}

\begin{figure}[ht]
     \centering
     \begin{subfigure}[b]{0.7\textwidth}
         \centering
         \includegraphics[width=\textwidth]{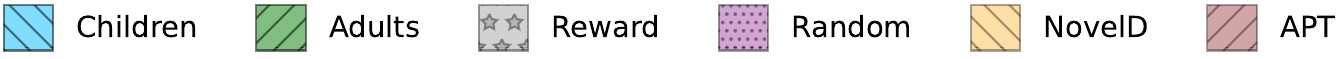}
         \label{fig:hist_legend_2}
     \end{subfigure}\\\vspace{-1\baselineskip}
     \begin{subfigure}[b]{0.3\textwidth}
         \centering
         \includegraphics[width=1.05\textwidth]{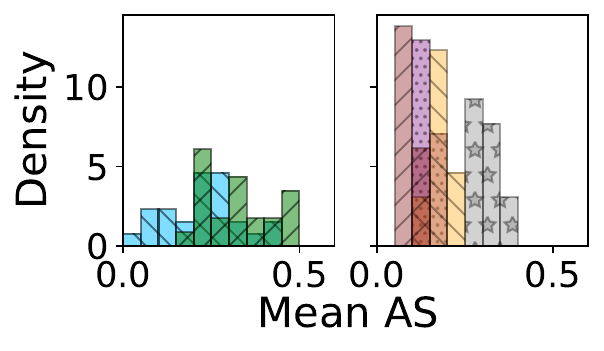}
         \label{fig:ep_achievement_score}
     \end{subfigure}
     \hfill
     \begin{subfigure}[b]{0.3\textwidth}
         \centering
         \includegraphics[width=1.05\textwidth]{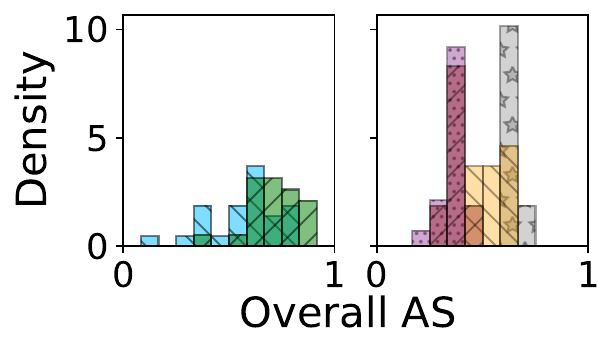}
         \label{fig:y equals x}
     \end{subfigure}
     \hfill
     \begin{subfigure}[b]{0.3\textwidth}
         \centering
         \includegraphics[width=\textwidth]{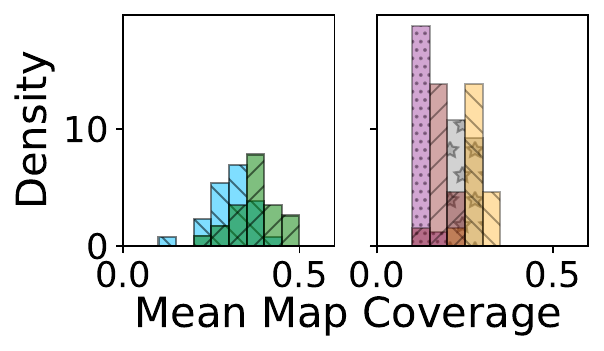}
         \label{fig:map_hist}
     \end{subfigure}
     \\\vspace{-1\baselineskip}
     \begin{subfigure}[b]{0.29\textwidth}
         \centering
         \includegraphics[width=1.05\textwidth]{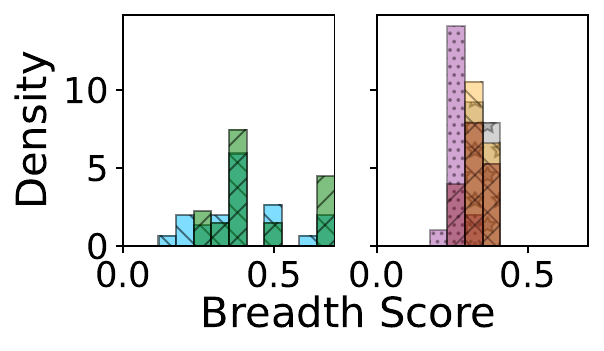}
         \label{fig:cu_breadth}
     \end{subfigure}
     \begin{subfigure}[b]{0.29\textwidth}
         \centering
         \includegraphics[width=1.05\textwidth]{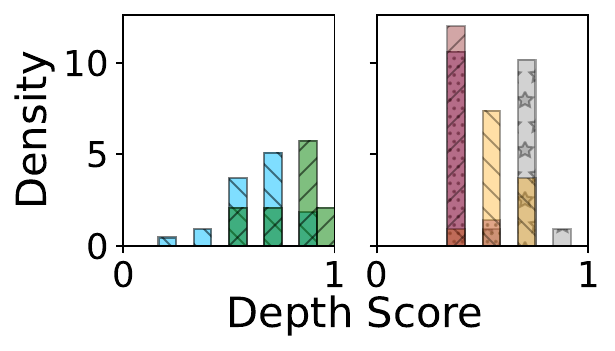}
         \label{fig:cu_depth}
     \end{subfigure}
     \begin{subfigure}[b]{0.38\textwidth}
         \centering
         \includegraphics[width=1.05\textwidth]{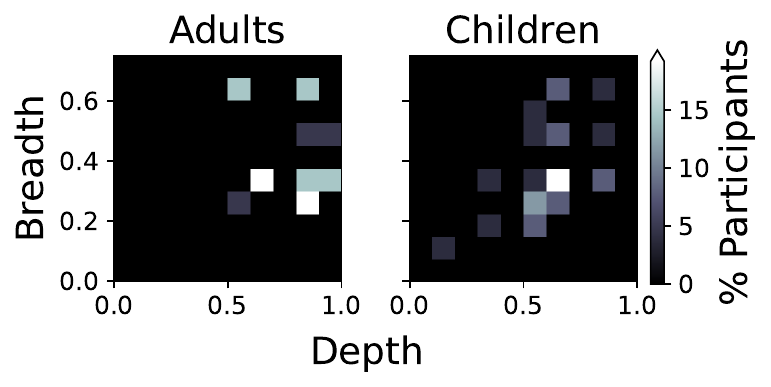}
         \label{fig:breadthvdepth}
     \end{subfigure}
     
     \vspace{-1\baselineskip}
        \caption{Summary density histograms for each exploration score. For clarity, the left plot for each measure shows human scores and the right plot shows agent scores.}
        \label{fig:summary-expl-hists}
\end{figure}

We present summary statistics across all human and agent data. First, Figure~\ref{fig:summary-expl-hists} shows the normalized density of humans and agents on the proposed exploration scores. To compare the distribution of scores in each group, we use Permutation tests on the means for the discrete Overall scores, and Wilcoxon Rank Sum tests for the continuous Mean AS and Map score, and find the following significant differences (p<0.05). Adults perform the best, scoring higher on average than children on all but Breadth, and higher than the intrinsic agents across all scores.
The extrinsic \textit{Reward} agent outperforms the intrinsic agents on most scores and even performs comparably to adults on the Mean Achievement Score---the score most similar to the extrinsic reward function, since both depend on the achievements reached in every episode. This demonstrates the impact of the reward function and suggests that there may be room for better intrinsic reward design to improve exploration. 

We also investigate whether different exploration strategies were present among the humans, and found that the Breadth and Depth scores were significantly positively correlated for the children ($\rho=0.51,p=0.01$) but not the adults ($\rho=-0.27,p=0.21$). The 2D histograms in Figure~\ref{fig:summary-expl-hists} show that the correlation between breadth and depth in the children is most apparent for the lower-scoring children, suggesting that it arises from a common cause of poor exploration ability. Different exploration styles become apparent in more proficient players, some of whom prefer to build on existing progress while others prefer to make more diverse independent discoveries.

\subsection{Information-Theoretic Intrinsic Objectives}
    
We investigate to what extent human and agent exploration optimize information-theoretic intrinsic objectives. We focus on \textit{Entropy}, \textit{Information Gain}, and \textit{Empowerment}, each of which have been proposed as exploration objectives in both the AI and cognitive science literature. 

\textbf{Entropy}
 measures the diversity of all the states visited throughout gameplay. We compute state visitation distribution $p(s) = N_s / \sum_{s'} N_{s'}$, where $N_s$ is the number of times the player was in state $s$. The entropy of this distribution is defined as 
\begin{equation}
    \textrm{H(\textit{S})} \coloneq \sum_{s \in S} -p(s)\log p(s).
\end{equation}
The entropy at episode $e$ is calculated over the visitation counts $N^e_s$ summed over all episodes up to and including $e$. The overall \textit{Entropy} is reported over each person or agent's total experience, summing counts across all episodes. 

\textbf{Information Gain}
 is a measure of how much information is gained on average over each episode, in terms of the novelty of the observed transitions. The total information accumulated up to the end of episode $e$ from taking action $a$ from state $s$ a total of $N^e_{s,a}$ times is measured as:
\begin{equation}
    IG_0^e(s,a) \coloneq \log(1+N^e_{s,a}).
\end{equation}
We use the log-count to approximate the diminishing novelty of repeated observations, as prior work has found it to perform similarly to more complex measures \citep{matusch2020evaluating}. The information gained just in episode $e$ is the difference in total information between that episode and the previous episode:
\begin{equation}
    IG^e(s,a) = IG_0^e(s,a)-IG_0^{e-1}(s,a).
\end{equation}
This is summed over all $(s,a)$ pairs and divided by the total number of transitions occurring that episode to get the average information gain per transition at episode $e$:
\begin{equation}
    IG^e \coloneq \frac{\sum_{(s,a)} IG^e(s,a)}{\sum_{(s,a)} N^e_{(s,a)}-N^{e-1}_{(s,a)}}.
\end{equation}
The overall \textit{Information Gain} is finally calculated as the mean of $IG^e$ over all episodes.

\textbf{Empowerment}
 can be interpreted as a measure of the amount of influence the agent has over the environment. The empowerment at state $s$ is defined as the channel capacity between the agent's actuators and the next state from $s$ \citep{salge2014empowerment}:
\begin{equation}
    \mathfrak{E}(s) \coloneq \max_{p(a)} I(S';A|s) =   \max_{p(a)} \sum_{a\in\mathcal{A},s'\in\mathcal{S}} p(s'|a,s)p(a) \log{\frac{p(s'|a,s)}{\sum_{\mathcal{A}}p(s'|a,s)p(a)}},
\end{equation}
where $I$ is mutual information. In other words, the empowerment at a particular state is given by the maximum mutual information between actions $a$ taken from that state and subsequent states $s'$.
For each participant we compute transition probabilities $p(s'|a,s)$ using observed transition counts $N_{s,a,s'}$ accumulated across all episodes. Next state distributions from unexplored states and actions are treated as uniform (corresponding to low empowerment), thus this measure can be viewed as a player's \textit{experienced} empowerment. We only count single step transitions $(s_t,a_t,s_{t+1})$ because most meaningful transitions in Crafter occur over one step. We use the Blahut-Arimoto \citep{dupuis2004blahut} algorithm to approximate the channel capacity, as proposed in \cite{klyubin2005all}, and sum over all states to obtain the total empowerment as
\begin{equation}
    \mathfrak{E}=\sum_{s \in \mathcal{S}} \mathfrak{E}(s).
\end{equation}

To tackle sparsity in our data, we use a reduced state representation to collapse the most semantically similar states. We combine the semantic label of the cell the player is currently facing, the contents of their inventory, and the increase in their status from the previous state, if any. This captures aspects of the environment that the player is most likely to be paying attention to and has direct control over, while aiming to avoid meaningless sources of variation in the state (\textit{e.g.}, slightly different positions of trees in the procedurally generated map). We use this representation to construct transition tables from the collected gameplay data, mapping each transition $(s,a,s')$, i.e., state, action, next state, to the number of times it was experienced. We further filter out movement transitions, focusing on transitions involving attempts to influence the environment.

\begin{figure}[ht]
     \centering
     \begin{subfigure}[b]{0.7\textwidth}
         \centering
         \includegraphics[width=\textwidth]{figs/histogram_horizontal_legend_children.pdf}
         \label{fig:hist_legend}
     \end{subfigure}\\\vspace{-1\baselineskip}
    \begin{subfigure}[b]{0.3\textwidth}
      \centering
      \includegraphics[width=1.05\textwidth]{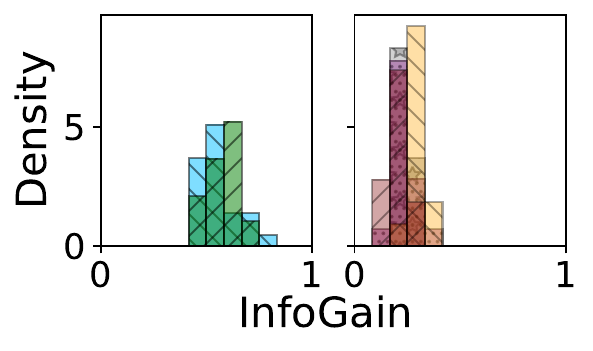}
    \end{subfigure} 
    \hfill
     \begin{subfigure}[b]{0.3\textwidth}
         \centering
         \includegraphics[width=1.05\textwidth]{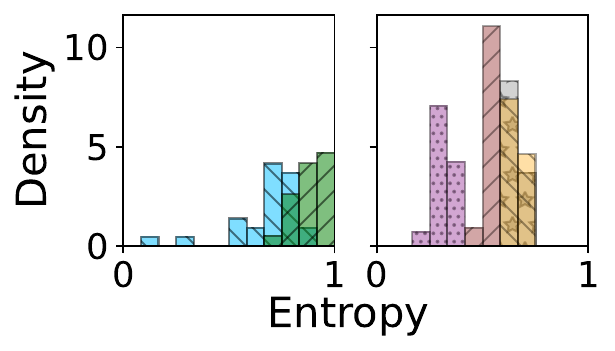}
     \end{subfigure}
     \hfill
     \begin{subfigure}[b]{0.3\textwidth}
         \centering
         \includegraphics[width=1.05\textwidth]{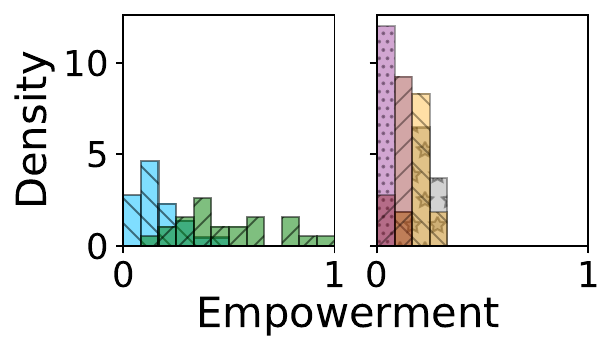}
     \end{subfigure}\\
    \begin{subfigure}[b]{0.75\textwidth}
      \centering
      \includegraphics[width=\textwidth]{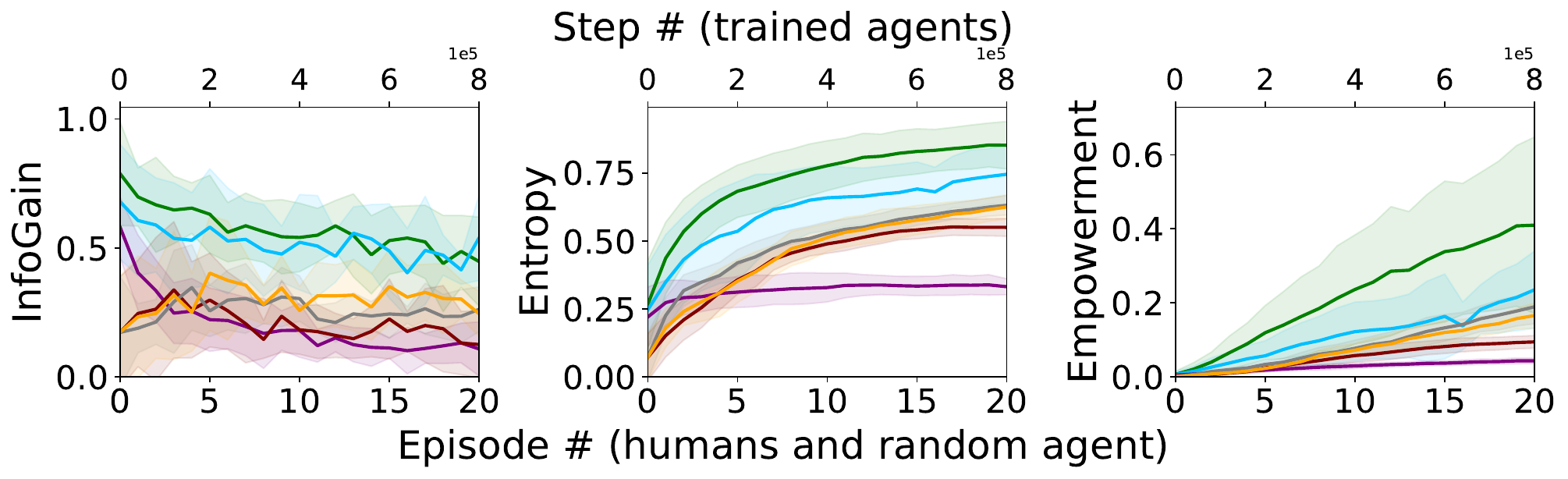}
    \end{subfigure} 
    \begin{subfigure}[b]{0.24\textwidth}
      \centering
      \includegraphics[width=\textwidth]{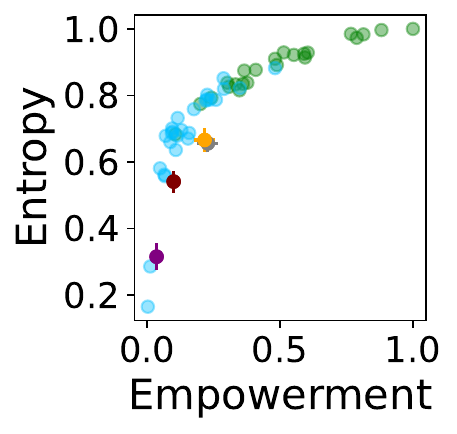}
    \end{subfigure} 
        \caption{Histograms showing the distribution of each normalized information-theoretic objective attained by each group of humans and agents, their means and standard deviations over time, and a scatter plot of overall Entropy and Empowerment for each human participant and the mean and standard deviations for each type of agent.}
        \label{fig:summary-info-hists}
\end{figure}

\textbf{Results.} Running Wilcoxon Rank-Sum tests on the intrinsic objectives, we find the following significant differences at $p<0.05$. Adults achieve higher levels of the objectives than all agents and children, except children's \textit{Information Gain} is on par with adults. Children also achieve higher or comparable \textit{Entropy} to the agents. NovelD and Reward display comparable \textit{Entropy} and \textit{Empowerment}, and both outperform children on \textit{Empowerment}, but NovelD achieves higher \textit{Information Gain}. This might be expected due to NovelD's information gain-inspired reward, but APT surprisingly performs worse than the other RL agents on all information-theoretic measures, even \textit{Entropy}.

Plotting the behavior over time (Fig. \ref{fig:summary-info-hists}), we observe that the \textit{Information Gain} per episode decreases for the humans as the most accessible observations get exhausted, but both adults and children maintain it at a significantly higher level than all the agents throughout their learning. All groups increase their \textit{Entropy} and \textit{Empowerment} throughout learning, though \textit{Entropy} increases more rapidly early on before plateauing, whereas \textit{Empowerment} increases at a steady rate. We also find overall \textit{Entropy} and \textit{Empowerment} exhibit a nonlinear positive correlation, where Empowerment increases more steeply for players that attain high Entropy. These results suggests that \textit{Entropy} could be a predominant objective earlier in exploration while there are still many novel states to observe, and then \textit{Empowerment} provides more signal once most states have been visited but there is still more to learn about how to predictably influence them.

\begin{figure}[ht]
     \centering
     \begin{subfigure}[b]{\textwidth}
         \centering
         \includegraphics[width=0.8\textwidth]{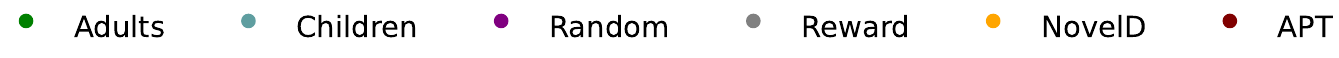}
         \label{fig:info_theoretic_legend}
     \end{subfigure}
     \vspace{-.0\baselineskip}
     \begin{subfigure}[b]{\textwidth}
         \centering
         \includegraphics[width=\textwidth]{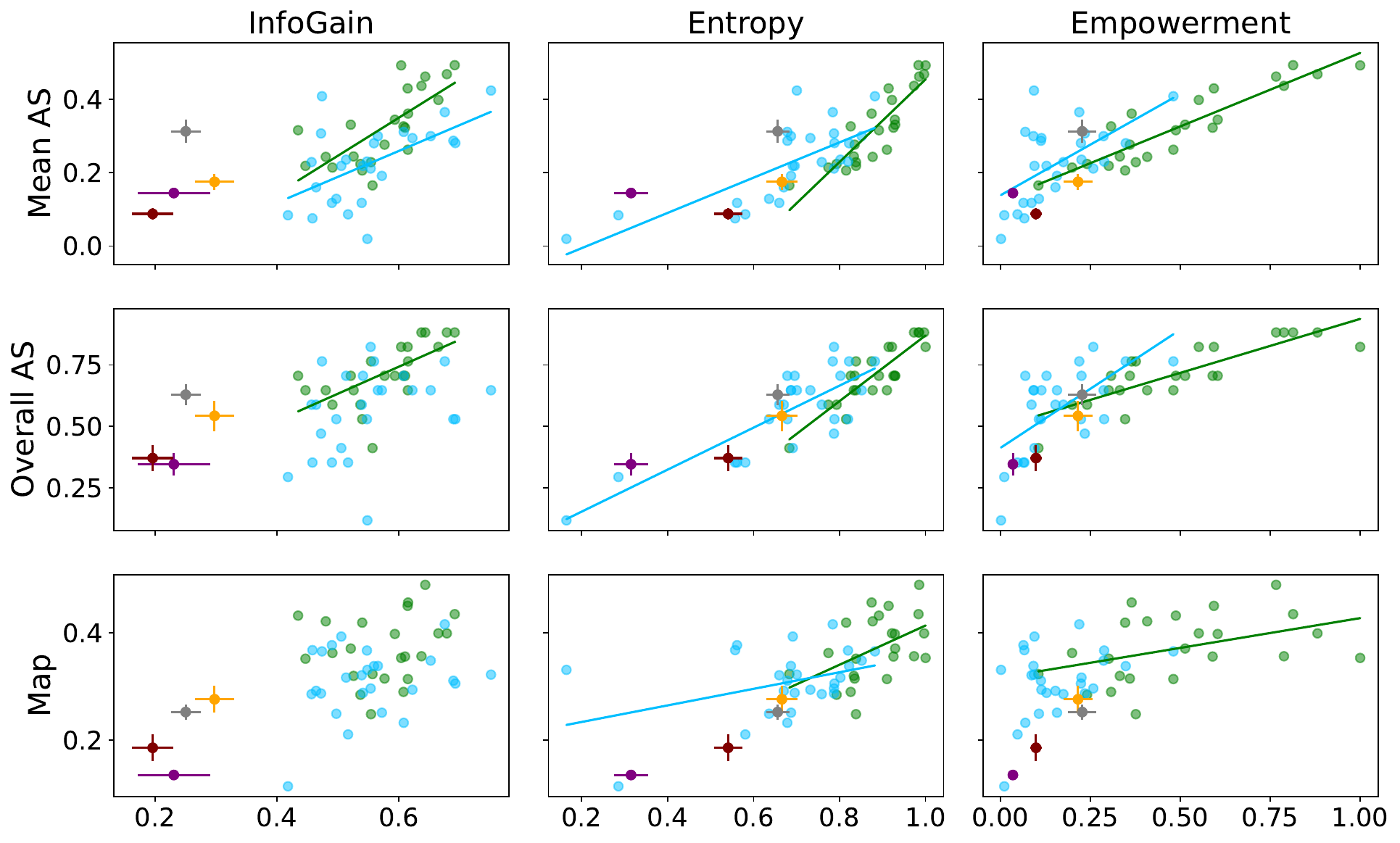}
         \label{fig:objectives_vs_scores}
     \end{subfigure}
     \caption{Information-theoretic objectives vs. exploration scores. Adults and children are scattered individually, while the mean and standard deviation across the random seeds is plot for each type of AI agent. The line of best fit is plot for adults and children when correlation is significant ($p<0.05$).}
     \label{fig:infovscore}
    \end{figure}

In Figure \ref{fig:infovscore}, we scatter the information-theoretic objectives against the exploration scores of each human participant and compute least squares linear fits for the adults and children separately, plotting it where significant ($p < 0.05$). We find significant positive correlations between humans' achievement-based exploration scores and their \textit{Entropy} and \textit{Empowerment}. \textit{Information Gain} may correlate less well with exploration success because it is possible to achieve high information gain by trying many different actions at every state, even though most actions have no effect.

\subsection{Private Speech} 
\begin{figure}[ht]
     \centering
    \begin{subfigure}[b]{0.374\textwidth}
      \centering
      \includegraphics[width=\textwidth,height=2.8cm]{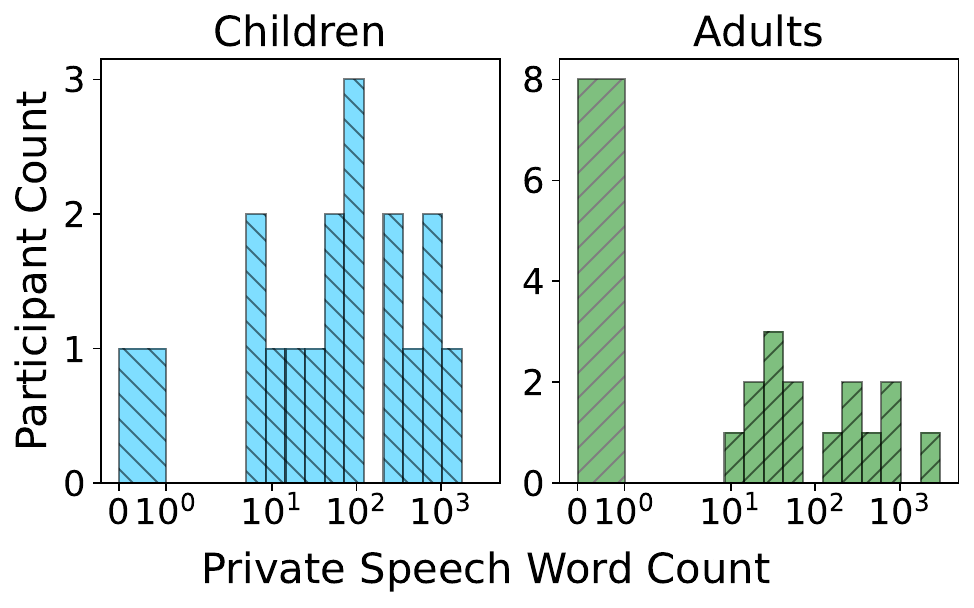}
    \end{subfigure} 
     \begin{subfigure}[b]{0.375\textwidth}
         \centering
    \includegraphics[width=\textwidth,height=2.8cm]{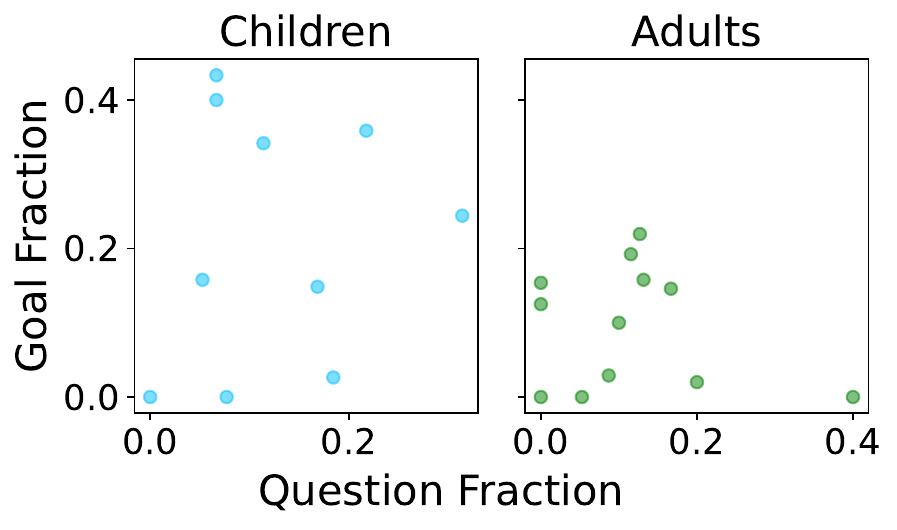}
     \end{subfigure}
     \begin{subfigure}[b]{0.2\textwidth}
         \centering
    \includegraphics[width=\textwidth,height=2.8cm]{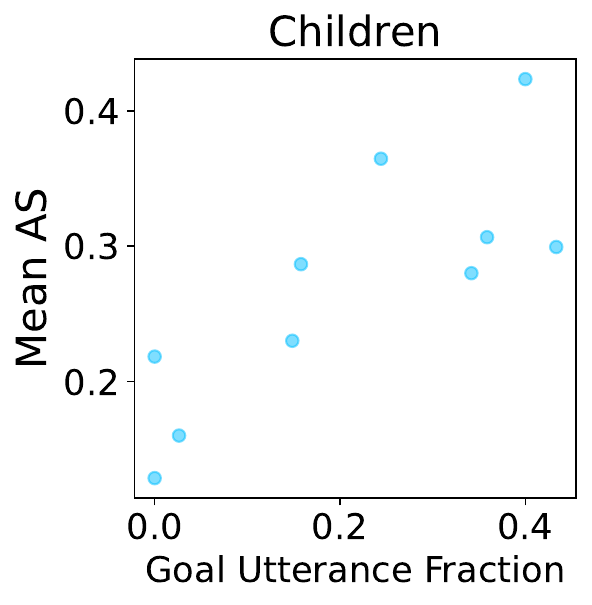}
     \end{subfigure}
        \caption{Left: the distribution of word counts from the subjects' private speech. Center: the fraction of private speech utterances that were classified as expressing goals versus questions. Right: the fraction of verbalized goals vs Mean Achievement Score in children. We find a strong significant correlation ($\rho=0.8, p=0.005$ unadjusted). No significant correlation was found for the fraction of questions, or for the adults with any scores.}
        \label{fig:private-speech}
\end{figure}

We present a preliminary analysis of the relationship between private speech and exploration. We take transcriptions from each consenting participant and filter for private speech utterances. Though many participants did not talk at all, a few uttered over 1000 words of private speech (see Figure~\ref{fig:private-speech}, left). We find that children spoke to themselves almost twice as much as adults, with an average word rate of about 16 words per minute compared to adults' approximately 8 words/minute, although both distributions had long tails. 

We find positive correlations between the children's word rate and their Mean Achievement Scores, \textit{Entropy}, and \textit{Empowerment}, although only significant at $p<0.05$ for \textit{Empowerment} (see Appendix~\ref{fig:private-speech-v-EE} for plots). Surprisingly, we also find negative correlations between the adults' word rates and all their achievement-based scores and information-theoretic objectives (though only significant at $p<0.05$ for \textit{Entropy}). 

It is possible that an increased rate of private speech in children simply reflects increased engagement with the game. To further investigate whether private speech actually helps children explore by helping them focus on achieving their goals, we investigated whether children who were more likely to verbalize goals performed better. 
We also analyzed how often children verbalized questions as a contrasting measure (see Table~\ref{tab:verbdefs} for examples).

\begin{table}[]
    \begin{tabular}{c|c}
             \toprule 
            Utterance type & Examples  \\ \midrule
            Questions & \parbox{0.7\linewidth}{``How do I make a wooden pickaxe?", ``What does sleep mean?", ``What is that skeleton doing?", ``Can I go in water?", ``where is stone?"}\\ 
            \midrule
            Goals &\parbox{0.7\linewidth}{``I need to mine some stuff", ``I wanna get the tree",  ``I need meat!", ``gonna kill that stuff.", ``now I can get some stones."} \\ \bottomrule
        \end{tabular}
        \caption{Examples of private speech utterances.}
        \label{tab:verbdefs}
\end{table}

Following prior works using LLMs for summarizing human data \citep{rathje2024gpt}, we take transcriptions from each child participant and use ChatGPT (\texttt{gpt-3.5-turbo}) to classify if each utterance expresses a goal and/or question. To improve accuracy, we also ask the LLM to generate reasoning before making each classification. See Appendix ~\ref{appendix:prompts} for sample prompts and responses. To reduce the effects of noise, we limit our analysis to participants who made at least 5 utterances throughout their gameplay, leaving us with N=10 children and N=12 adults. 

We find that while the fraction of question utterances was almost the same for children and adults at 12-13\%, children were on average more than twice as likely to express goals than adults, at 21\% versus 10\% (see Figure~\ref{fig:private-speech} center). And while we find no correlation between the fraction of questions and any measures of exploration performance, we find a striking positive correlation ($\rho=0.8, p=0.005$) between the fraction of goal utterances and the children's Mean Achievement Score (see Figure~\ref{fig:private-speech}).

\section{Discussion}

\textbf{Conclusions.}
In this work we investigate the role that information-theoretic intrinsic motivation may play in driving open-ended exploration in complex environments. We find evidence that both \textit{Entropy} and \textit{Empowerment} are consistently positively correlated with exploration progress. However, agents trained on intrinsic rewards explore less effectively than adults, and fail to attain higher \textit{Entropy} or \textit{Empowerment} than even agents trained on extrinsic rewards focused on the task structure in the environment.
This suggests that intrinsically-motivated agent exploration could be improved through better ways of approximating and incorporating entropy and empowerment-based objectives. Furthermore, we observe that \textit{Entropy} increases rapidly before plateauing and \textit{Empowerment} increases slowly and steadily throughout gameplay, which suggests that they may be even more effective when combined: \textit{Entropy} could guide early exploration while \textit{Empowerment} may provide more signal for advanced exploration. 

Finally, although the children under-perform the adults on most scores and objectives except \textit{Information Gain}, we find preliminary evidence suggesting their performance may be aided by overt verbalization. Children with higher word rates perform better, and furthermore we find a positive correlation between the fraction of utterances specifically mentioning \textit{goals} and the rate of unlocking achievements.  These results are consistent with prior findings in psychology that self-talk can help direct and focus problem solving in children, and findings in AI that agents that generate goals with language may explore more effectively \citep{Lee_2011,zhang2023omni,du2023guiding}.

\textbf{Limitations.} 
The small sample size limits broader conclusions about human exploration in different settings and populations. We also note that the analyses on verbalizations were exploratory, and need to be confirmed with a larger sample in a preregistered study format. Our calculations of the information-theoretic objectives are sensitive to the state representation used; although we tried to mimic human prior knowledge by using a relatively simple and compact semantic representation, humans' priors may allow them to compress the state space in even more effective ways. That said, we hope this work inspires interest in the intersection between cognitive science and AI, laying ground for future work that can collect larger datasets in richer and more naturalistic settings. 
\section{Methods}\label{sec:methods}

\subsection{Open-Ended Environment: Crafter} 
\begin{figure}
    \centering
    \includegraphics[width=\linewidth]{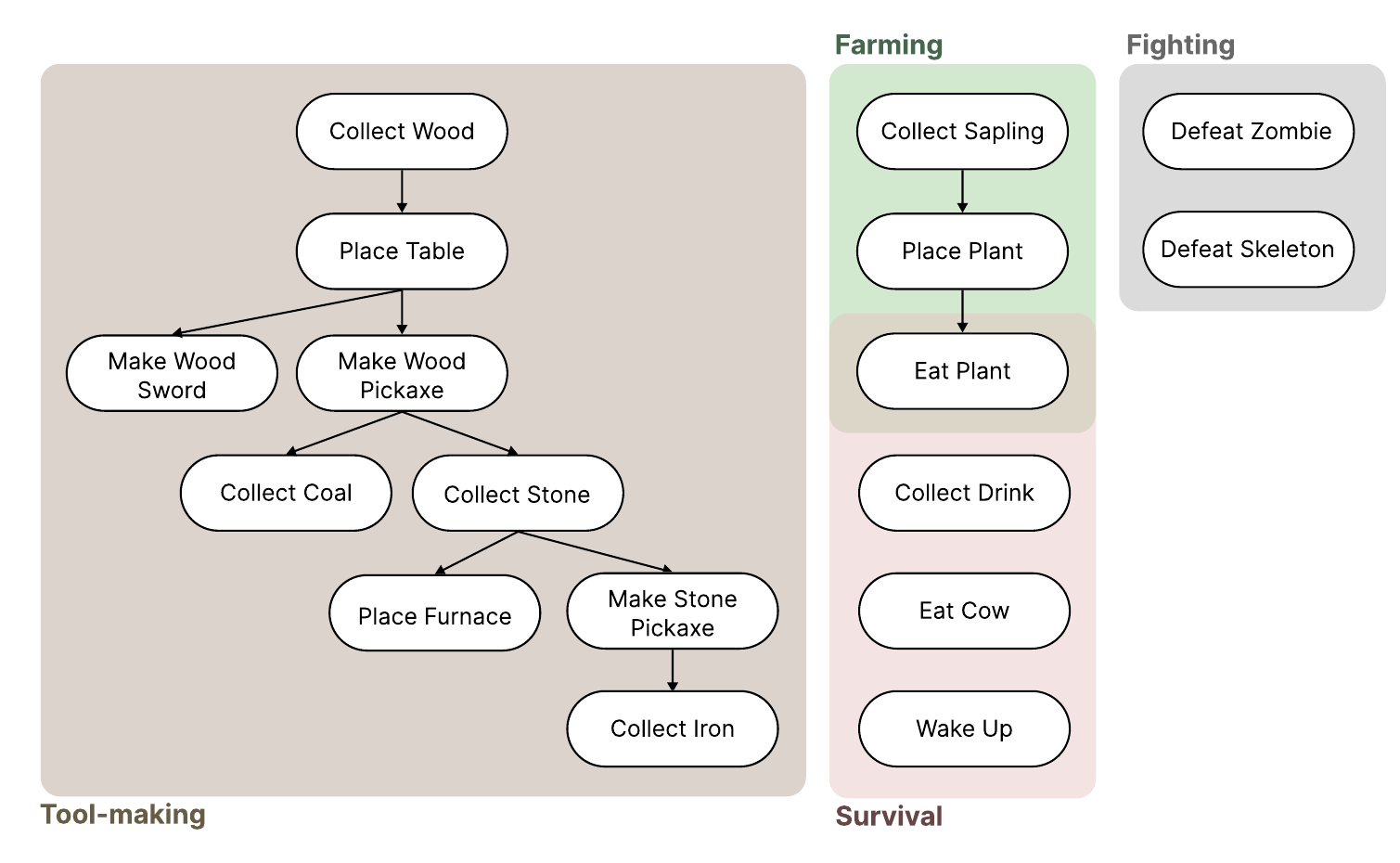}
    \caption{Achievements that can be unlocked in our modified version of Crafter, roughly clustered into four main categories (modified from Figure 4 in \cite{hafner2021benchmarking}). This is for illustrative purposes; no information about these achievements is exposed to the human participants or the intrinsically motivated RL agents.}
    \label{fig:tasks}
\end{figure}

We conduct our experiments in Crafter \citep{hafner2021benchmarking}. Similar to Minecraft, Crafter allows the player to explore a wide breadth and depth of activities. The player controls a character in a procedurally generated grid-world containing various resources that can be collected and used to replenish status levels (food, water, energy) or build tools (Figure \ref{fig:screen}). Health status decreases when the player is attacked or any of the other status levels are at 0. The episode ends when the health reaches 0, resetting the player into a new randomly-generated world with an empty inventory. All semantically meaningful changes in state that are under the player's control are marked by \textit{achievements}, some of which have multi-level prerequisites that must be satisfied within the same episode (see Figure \ref{fig:tasks}).

The available actions either move the player or interact with the environment. Interactions only affect the cell that the player is directly facing, with the "do" action being the most versatile (used to eat, drink, cut trees, mine, and fight). Seven additional actions each have unique effects on the environment, sometimes requiring prerequisites to be satisfied (\textit{e.g.,} \texttt{place crafting table} only works if the player has sufficient wood in their inventory). 

We modify the game to make it easier to master in a short play session and accessible via a conventional game controller. First, we slightly lengthen the fraction of an in-game day that is spent in daylight by changing the daylight function from $1-\left|\cos\left(\pi x\right)\right|^{3}$ to $1-\left|\cos\left(\pi x\right)\right|^{12}$. We also add an explicit `Game Over' screen when an episode ends to ensure participants are aware of when their character dies. Lastly, we slightly prune the action space, removing five of the most advanced actions from the original game as they were not feasible to fit on a conventional handheld game controller (see Figure \ref{fig:screen}, center).

\subsection{Participants} In this pre-registered, IRB approved study (AsPredicted reference: 92521), we recruited 51 children between the ages of 6-12 years from the Bay Area Discovery Museum (BADM), as well as 24 adults from the University of California, Berkeley) campus ages 18-25 years (Mean age 24.8, Female: 10, Male: 14). Participants who were not able to complete at least one full episode of the game, or who received significant verbal interference from onlookers at the BADM, were excluded. Preliminary evaluations suggested that the game was too challenging for younger children, so we only analyzed data from children at least 8 years old, resulting in data from 26 children (Mean age: 9.4, Female: 11, Male: 15). We found that 92\% of children had video game experience with 77\% having Minecraft-specific experience, and 79\% of adults had video game experience with 54\% having played Minecraft previously.

\subsection{Data Collection Procedure} Participants were first shown a short tutorial video explaining what each controller button did (Figure \ref{fig:screen}), without specifying any scoring or objective so we can observe intrinsically-motivated behavior. 
Participants were then allowed to play for up to 20 minutes, with the option to quit early. All actions taken and the complete world state was recorded for every time-step while playing. In addition, audio from participants who consented to audio recording was later transcribed manually with timestamps. This resulted in verbal transcripts from 17 children and 23 adults.

\subsection{Agent Training Procedure}
We trained three RL agents and used one random agent as a baseline. At each timestep, the random agent takes no action with probability 0.475 in order to match the average action rate of the human players, and uniformly samples all available actions otherwise. We trained two agents (\textit{NovelD} and \textit{APT}) with state-of-the-art intrinsic RL objectives. NovelD~\citep{zhang2021noveld} incentivizes information gain by providing a large intrinsic reward at the boundary between familiar and novel regions, using RND \citep{burda2018exploration} as a measure of state novelty. APT~\citep{liu2021behavior} uses a particle-based entropy estimator \citep{singh2003nearest} to reward the agent for maximizing state entropy in an abstract representation space. As a measure of best-case performance we also train an extrinsic agent (\textit{Reward}) with a hand-crafted reward function that encourages it to stay alive and unlock in-game achievements (Fig.~\ref{fig:tasks}) -- which are also the basis for most of our exploration performance metrics. This reward function provides a sparse reward of 1 every time an achievement is reached for the first time in each episode, alongside a small reward of $\pm 0.1$ every time the agent replenishes or loses health. The agent policy input is a simplified semantic representation of the game: the labels of the contents in each cell in the agent's 9x8 viewing window, the status, and the inventory (Fig.~\ref{fig:screen}, left). All RL agents are trained with Rainbow DQN \citep{hessel2018rainbow} for one million timesteps, with $\epsilon$-greedy exploration decaying $\epsilon$ from 1 to 0.01 on a linear schedule over 250,000 timesteps. We train each agent with 13 random initializations. Finally, for our behavioral analysis we subsample 25 episodes evenly throughout the training trajectory of each agent to reduce the number of episodes to the same order of magnitude as the humans (which ranged from 5-28).

\subsection{Dataset and Code Availability} Our dataset of game-play and transcripts is available at \url{https://github.com/alyd/humans_in_crafter_data}, and our analysis code will be released upon publication. We hope it will be a useful resource for future research.
\backmatter


\section*{Acknowledgements}
We are grateful to Danijar Hafner, Julian Jara-Ettinger, Alane Suhr, Steve Piantadosi, Terry Regier and the participants of the MURI Group for discussion, and to the participants and their parents, the museums and preschools who participated, and the undergraduates who assisted, Bryanna Kauffman, Zane Levin, Elijah Phipps, Janie Dent, Jenna Levin, Nikita Kumar and Athena Leong. 

Funding: we gratefully acknowledge the support of the following agencies: DOD ONR MURI Self-Learning Perception Through Real World Interaction (grant number N00014-22-1-2773), the Center for Human-Compatible AI (CHAI) and the Fannie and John Hertz Foundation.

\section*{Declarations}

\begin{itemize}
\item Ethics approval and consent to participate: All studies were had full International Board Approval under the Gopnik Lab  and required full consent to participate.
\item Consent for publication: All participants consented and all children's names have been voided from any data published. Some consented to partial or full data publication. Initial Data was preregistered under AsPredicted reference: 92521.
\item Author contribution: 
Aly Lidayan contributed to the AI agent models, behavior and speech analysis,  and manuscript writing. 
Yuqing Du contributed to experimental design, gathering data in adults, the AI agent models, data analysis and manuscript writing. 
Eliza Kosoy contributed to experimental design, gathering data in children and adults, data analysis and manuscript writing. 
Maria Rufova contributed to data collection and curation, and speech analysis. 
Pieter Abbeel contributed to experimental design and thorough discussions on data analysis and results.
Alison Gopnik contributed to experimental design, thorough discussions on data analysis and results as well as manuscript editing.
\end{itemize}






\begin{appendices}

\newpage
\section{Supplementary Data}\label{secA1}

\subsection{Exploration Trajectories Over Time}

\begin{figure*}[ht]
\begin{center}
\includegraphics[width=0.6\textwidth]{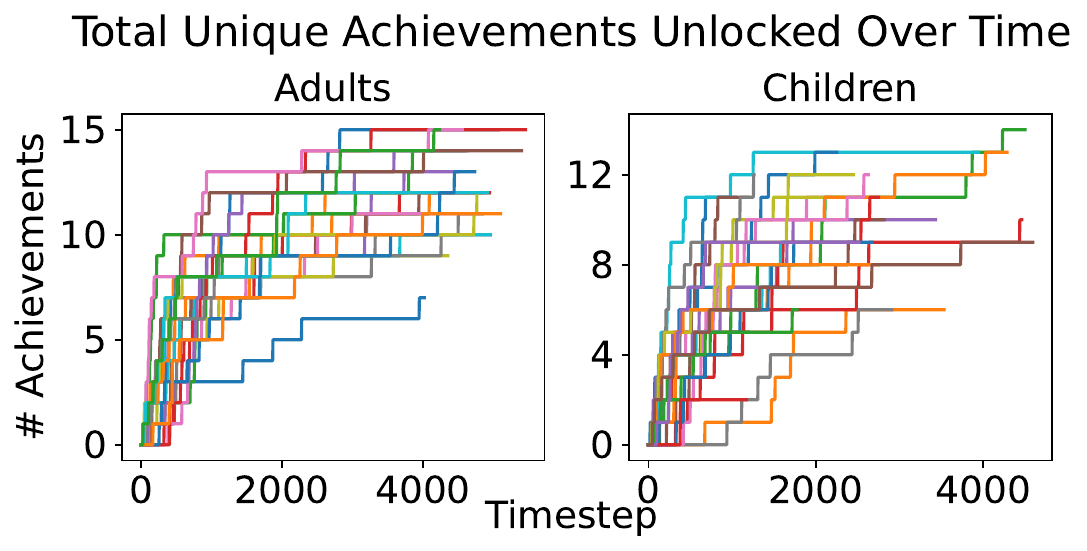}
\end{center}
\caption{Achievement unlocking trajectories for humans, concatenating all episodes together in chronological order.} 
\label{fig:unlocking}
\end{figure*}

\begin{figure}[ht]
\begin{center}
\vspace{-3ex}
\begin{subfigure}[b]{0.7\textwidth}
         \centering
         \includegraphics[width=\textwidth]{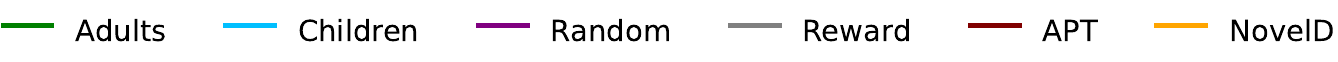}
     \end{subfigure}\\
\begin{subfigure}[b]{\textwidth}
         \centering
         \includegraphics[width=\textwidth]{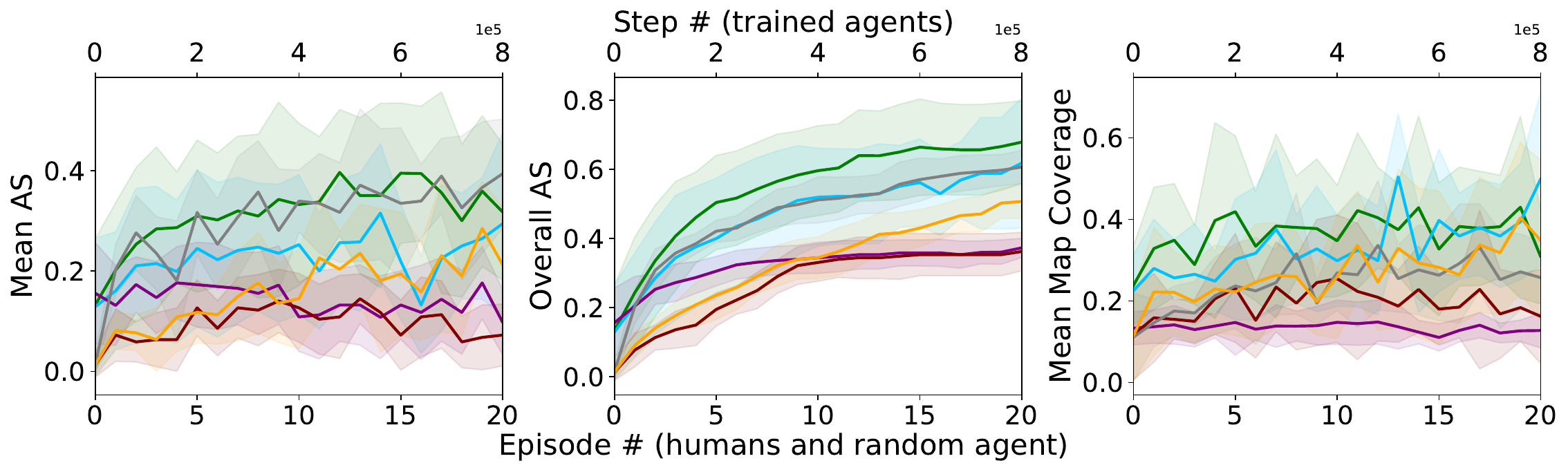}
     \end{subfigure}
 \\\vspace{-1\baselineskip}
     \begin{subfigure}[b]{\textwidth}
         \centering
         \includegraphics[width=0.666\textwidth]{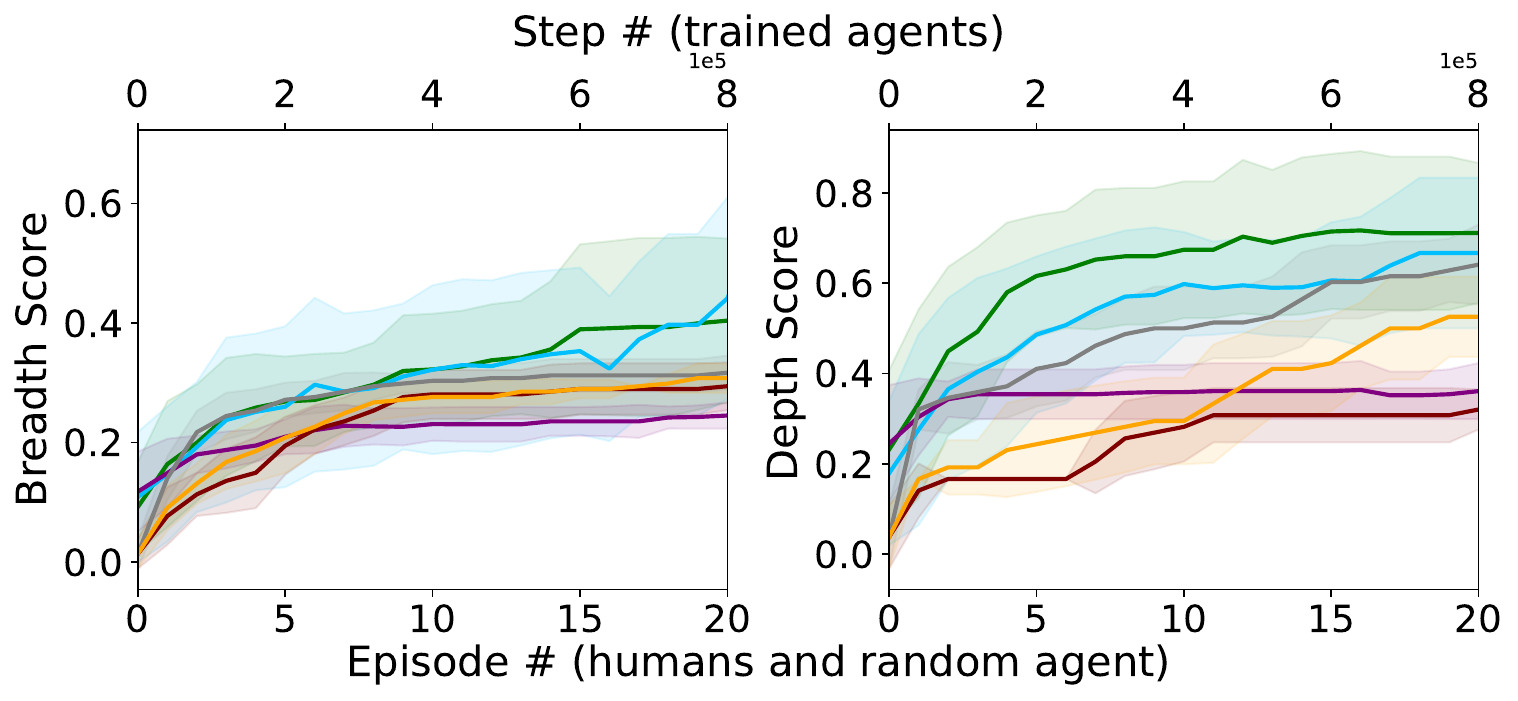}
     \end{subfigure}
\end{center}
\caption{The evolution of exploration scores over time. Agents are much less sample efficient than humans, reaching similar performance only after over 100$\times$ the number of environment interactions used by humans.} 
 \label{fig:scores_time}
\end{figure}

We take a random subset of the children and adults and examine their exploration trajectories individually, by plotting how the total number of unique achievements they have ever unlocked increases over time (see figure \ref{fig:unlocking}). 

We find that achievements are often unlocked at a very high rate in the first 500 time-steps (i.e. approximately first 100 seconds of gameplay). This is largely due to the fact that there are some easy achievements that can be unlocked very quickly, and the achievements further down the dependency tree take much longer. After this, players tend to unlock achievements at a slower but still steady rate. Again, adults on average do better than children- and there are far more children who quit playing early, even including some who were making very rapid progress.

\subsection{Effect of Prior Minecraft Experience}\label{appendix:mcxp}

\begin{figure}[ht]
\begin{center}
\begin{subfigure}[b]{\textwidth}
         \centering
         \includegraphics[width=\textwidth]{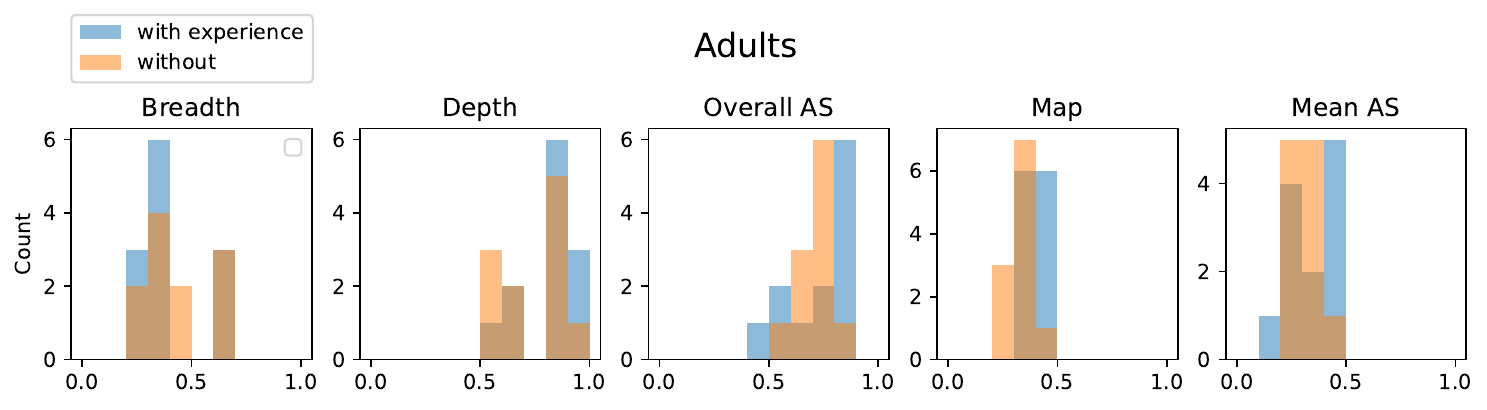}
     \end{subfigure}
 \\\vspace{-1\baselineskip}
     \begin{subfigure}[b]{\textwidth}
         \centering
         \includegraphics[width=\textwidth]{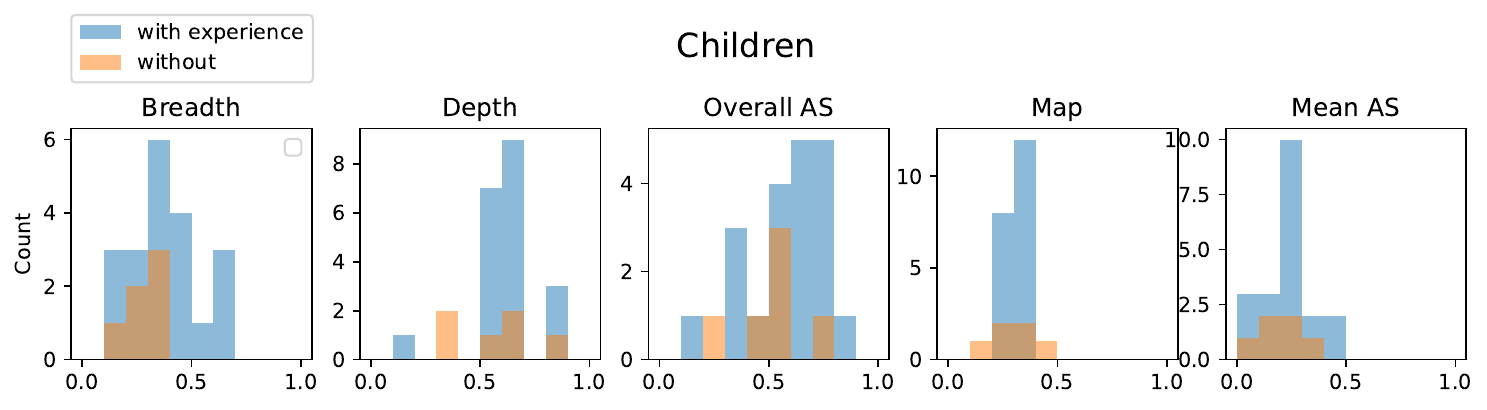}
     \end{subfigure}
     \\\vspace{-1\baselineskip}
     \begin{subfigure}[b]{\textwidth}
         \centering
        \includegraphics[width=\textwidth]{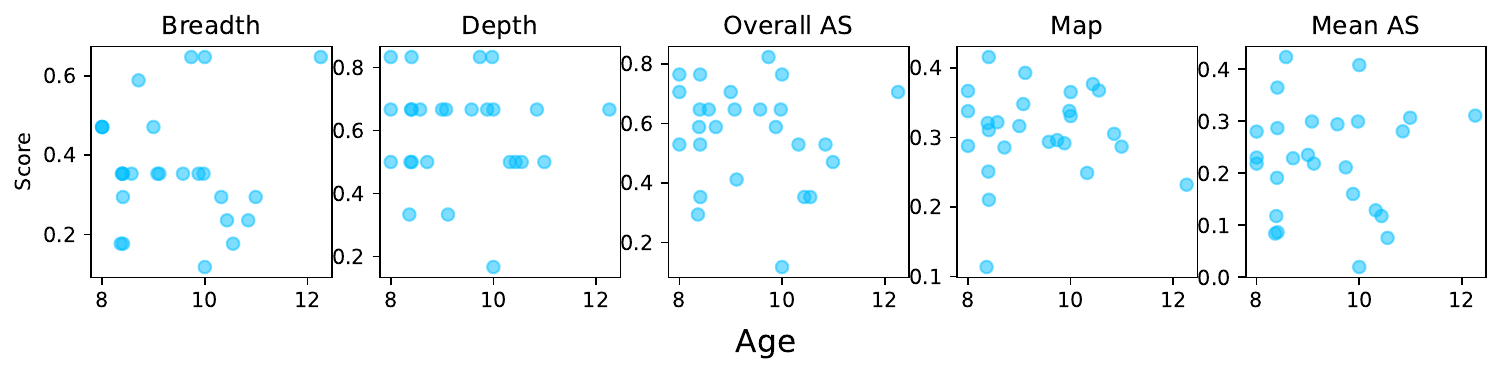}
     \end{subfigure}
\end{center}
\caption{The effect of prior Minecraft experience and age on performance.} 
 \label{fig:minecraft_exp}
\end{figure}

\newpage
\subsection{Prompting for Utterance Classification}\label{appendix:prompts}

Goal Prompt:
\begin{lstlisting}
 The following text is a part of a transcript from a person playing a game for the first time. Is the person expressing a goal? Provide reasoning and answer with Finish[1] for yes or Finish[0] for no or if it cannot be determined.\n EX: I need more woooOODD! A: This is a goal because the person is expressing their need for wood resources. Finish[1]\n EX: Wait can I even get, oh yea. A: This is not a goal because the person is stating an incomplete thought. Finish[0]\n EX:
\end{lstlisting}

Sample responses:

\begin{tabular}{c|c}
         \toprule 
        Input Utterance & Response  \\ \midrule
                
        \parbox{0.29\linewidth}{sooo, next thing build the table with your wood. also make the pickaxe.} &\parbox{0.7\linewidth}{A: This is a goal because the person is outlining specific tasks they want to accomplish in the game. Finish[1]} \\ \midrule
        \parbox{0.29\linewidth}{i should run, oh no, oh noooo. }& \parbox{0.7\linewidth}{A: This is not a clear goal as the person is expressing panic rather than a specific objective. Finish[0]}\\ 
        \midrule
        \parbox{0.29\linewidth}{um maybe i should just figure out what happens when i die }&\parbox{0.7\linewidth}{A: This is a goal because the person is expressing a desire to figure out what happens when they die in the game. Finish[1]} \\ 
        \midrule
        \parbox{0.29\linewidth}{which button to go left? red. }&\parbox{0.7\linewidth}{This is not a goal because the person is asking a question about game controls. Finish[0]} \\ 
        \bottomrule
    \end{tabular}

Question Prompt:
\begin{lstlisting}
The following text is a part of a transcript from a person playing a game for the first time. Is the person asking a question? Provide reasoning and answer with Finish[1] for yes or Finish[0] for no or if it cannot be determined.\n EX: why is the blue button not working? A: This is a question because the person is asking a question about the game. Finish[1]\n EX: Yes. A: This is not a question because it is just a yes or no answer. Finish[0]\n EX:
\end{lstlisting}

\begin{tabular}{c|c}
         \toprule 
        Input Utterance & Response  \\ \midrule
        \parbox{0.29\linewidth}{how much stone do i need for stone pickaxe again?}& \parbox{0.7\linewidth}{This is a question because the person is asking for information about the game. Finish[1]}\\ 
        \midrule
        \parbox{0.29\linewidth}{it’s hard, it’s hard. what? ready. it's not moving.}& \parbox{0.7\linewidth}{This is not a clear question as the person seems to be expressing frustration and confusion rather than asking a direct question. Finish[0]}\\ 
        \midrule
        \parbox{0.29\linewidth}{wait, can i hit this thing?}&\parbox{0.7\linewidth}{A: This is a question because the person is asking if they can interact with something in the game. Finish[1]} \\ 
        \midrule
        \parbox{0.29\linewidth}{i need to drink water. drink drink. }&\parbox{0.7\linewidth}{A: This is not a question, it is a statement. Finish[0]} \\ 
        \midrule
        \parbox{0.29\linewidth}{where is stone? great.} &\parbox{0.7\linewidth}{A: This is a question because the person is asking where something is in the game. Finish[1]} \\ \bottomrule
    \end{tabular}

\begin{figure}[ht]
     \centering
     \begin{subfigure}[b]{0.4\textwidth}
         \centering
         \includegraphics[width=1.03\textwidth]{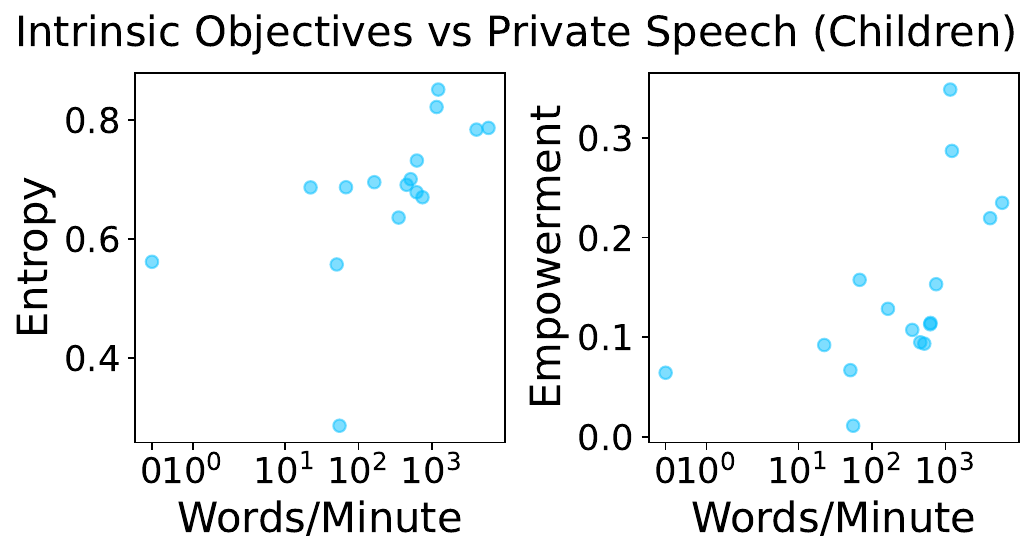}
     \end{subfigure}
      \hspace{0.05\textwidth} 
     \begin{subfigure}[b]{0.4\textwidth}
         \centering
         \includegraphics[width=1.03\textwidth]{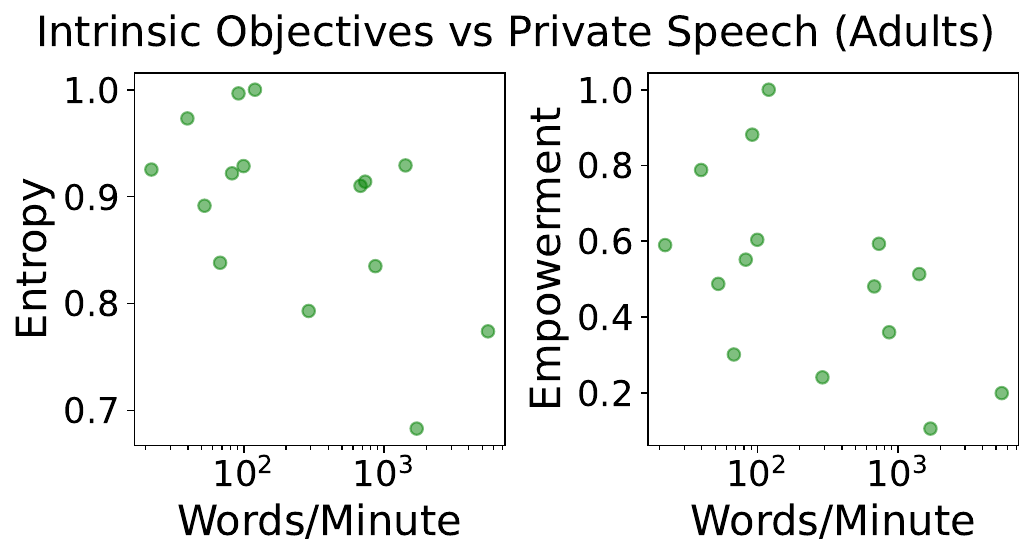}
     \end{subfigure}
        \caption{Measures of intrinsic objectives (entropy, empowerment) vs. words per minute of private speech, for Children and Adults.}
        \label{fig:private-speech-v-EE}
\end{figure}




\end{appendices}


\bibliography{sn-bibliography.bib}

\end{document}